\documentstyle[11pt]{article}
\title{Development of a \\
Spanish Version of the Xerox Tagger}
\author{\small \\
Fernando S\'anchez Le\'on \\
Laboratorio de Ling\"u\'{\i}stica Inform\'atica \\
Facultad de Filosof\'{\i}a y Letras \\
Universidad Aut\'onoma de Madrid \\
{\tt fsanchez@ccuam3.sdi.uam.es} \\
\\
Amalio F. Nieto Serrano \\
Departamento de Ingenier\'{\i}a de Sistemas Telem\'aticos \\
Escuela Superior de Ingenieros de Telecomuniaciones \\
Universidad Polit\'ecnica de Madrid \\
{\tt anieto@dit.upm.es} \\
\\
Doc. Id.: CRATER/WP6/FR1}
\date{May 19, 1995}

\begin{document}

\maketitle

\begin{abstract}
\small
This paper describes work performed withing the CRATER ({\em C}orpus
{\em R}esources {\em A}nd {\em T}erminology {\em E}xt{\em R}action,
MLAP-93/20) project, funded by the Commission of the European
Communities.  In particular, it addresses the issue of adapting the
Xerox Tagger to Spanish in order to tag the Spanish version of the ITU
(International Telecommunications Union) corpus.  The model implemented
by this tagger is briefly presented along with some modifications
performed on it in order to use some parameters not probabilistically
estimated.  Initial decisions, like the tagset, the lexicon and the
training corpus are also discussed.  Finally, results are presented and
the benefits of the {\em mixed model} justified.
\end{abstract}

\section{Introduction}

\normalsize
This paper describes the adaptation work carried out to retarget the
Xerox Tagger to Spanish\footnote{This work has been developed in the
context of the R\&D project CRATER ({\em C}orpus {\em R}esources {\em
A}nd {\em T}erminology {\em E}xt{\em R}action, MLAP-93/20), funded by
the Commission of the European Communities. Other partners involved in
the project are University of Lancater (UK), Computers,
Communications and Visions, C$^{2}$V (France) and IBM-France.}. The
Xerox Tagger \cite{Cutting92} has as one of its virtues the
characteristic of being based on a simple probabilistic model, as it
will become clear below. It is also claimed to be language-independent
and it is public domain\footnote{It can be obtained via {\tt ftp} from
parcftp.xerox.com under the directory pub/tagger. The program runs on
Common Lisp, and several implementations have been tested under SunOS
4.x and 5.x, besides that for the Macintosh.}. Various authors have
already developed ports to languages other than English (the language in
which development was performed). Thus, \cite{Feldweg95} presents the
issue of adapting the tagger to German, while \cite{Chanod95} report on
French. These ports have been performed at the same time to that
presented here for Spanish.

The interest on the Xerox Tagger not only comes from the virtues already
mentioned, but also benefits from the attraction that stochastic
approaches to Natural Language Processing have revived in the researchers
on the field. Widely commented examples of this resurgence of
probabilistic techniques include the double special issue that {\em
Computational Linguistics} devoted recently to this venture.
Nevertheless, it is more interesting in this debate the possibility to
combine (empirical and racionalist) techniques, rather than approaching
to statistical models with a ``let's-see-what-it-can-do'' idea.
Although they are capable of ``doing things'', with relative simplicity
in the estimation of parameters and in a robust way, being this
quality, as it is known, difficult to find in knowledge-based systems.

As a consequence of this combination of techniques, \cite{Tapanainen94}
present excelent results tagging an English corpus with the Xerox Tagger
and the constraint-based system ENGCG \cite{Karlsson94}. In this case,
the combination is performed by means of separate modules. In this
paper, and in a more modest way, a combination of techniques within the
Xerox Tagger is proposed.

\section{The Xerox Tagger}

The Xerox Tagger uses a statistical method for text tagging. In these
systems, ambiguity of assignment of a tag to a word is performed on the
basis of most likely interpretation. A form of Markov model is used
that assumes that a word depends probabilistically on just its
part-of-speech, which in turn depends, in most systems though not in
the Xerox Tagger, solely on the category of the preceding two words.

Two types of training have been used with this model. The first one
makes use of a tagged training corpus. A small amount of text is
manually tagged and used to train a partially accurate model. This
model is then used to tag more text; the tags are manually corrected
and subsequently used to retrain the model. This training method has
been called {\em bootstraping} \cite{Derouault86}.

The second method does not require a tagged training corpus. The model
is then called a {\em hidden Markov model} (HMM), as state transitions
cannot be determined while the sequence of outputs is known.
\cite{Jelinek85} uses this method for training a text tagger.  A
three-gram approach is generally used, where trigram estimates are
smoothed out using the method of {\em deleted interpolation} in which
weighted estimates are taken from second- and first-order models and a
uniform probability distribution.  \cite{Kupiec89b} uses word
equivalence classes based on parts of speech, to pool data from
individual words.  The most common words are still stored in a
lexicon file, while all other words are represented according to the
set of possible categories they can assume.  The number of equivalence
classes (referred to as {\em ambiguity classes} in \cite{Cutting92})
can be considerably reduced (to aprox.  400 for the whole vocabulary
contained in the Brown corpus).  As a further reduction of the number
of parameters, a first-order model can be employed.  In these models, a
word depends on its part-of-speech category, which depends solely on
the category of the preceeding word.  \\

The Xerox Tagger is based on an HMM. It uses ambiguity classes and a
first-order model to reduce the number of parameters to be estimated
without significant reduction in accuracy. According to the authors,
reasonable results can be produced training on as few as 3,000
sentences. Besides, ``relatively few ambiguity classes are sufficient
for wide coverage, so it is unlikely that adding new words to the
lexicon requires retraining, as their ambiguity classes are
accomodated.'' Words not found in the lexicon are assigned an ambiguity
class according to both context and {\em suffix} information\footnote{The
term {\em suffix} must be understood in this context in a wide sense
(set of ending characters in a word) and not strictly linguistic.}. \\

\subsection{Procedure}

Lets briefly describe the procedure of the tagger. After the {\em
tokenizer} has converted the input text into a sequence of {\em
tokens}, these tokens are passed to the {\em lexicon}. Tokens are
converted into a set of stems, each annotated with a part-of-speech
tag. The set of tags identifies an {\em ambiguity class}, which is also
delivered by the lexicon.

The {\em training} module takes long sequences of ambiguity classes as
input. It uses the Baum-Welch algorithm to produce a {\em trained HMM},
as input to the tagging module. The {\em tagging} module buffers
sequences of ambiguity classes between sentence boundaries. These
sequences are disambiguated by computing the maximal path through the
HMM with the Viterbi algorithm.

Words not found in the manually-constructed lexicon ``are generally both
open class and regularly inflected'', according to
\cite{Cutting92}\footnote{However, this greatly depends on the size of
the lexicon.  Since exhaustive lexicons are ``expensive, if not
impossible, to produce'', in authors' words, this statement may become
false.}.  A language-specific method can be employed to guess ambiguity
classes for these unknown words.  Hence, the Xerox Tagger provides a
function that computes `suffixes' together with probabilistic
predictions of a word's category ending in each of the suffixes
calculated.  This function also operates on an untagged training
corpus.

As a final stage, words not found in the lexicon and ending in a suffix
not recognized are assigned a default ambiguity class (open class). \\

\section{A mixed model}

As already mentioned in the previous section, in case a word is unknown
to the system, `suffix' information can be used in order to approximate
its possible ambiguity class. This information can be calculated by
means of the LISP function {\tt class-guesser:train-guesser-on-files}.
The authors strongly recommend the use of this function in order to
retarget the tagger to new corpora, new tagsets, and new languages [Jan
Pedersen, personal communication].  However, we will try to demonstrate
that a system using a set of manually-added suffixes performs better,
at least for inflectional languages like Spanish. \\

The above-mentioned function operates on a training text and calculates
two parameters:

\begin{itemize}

\item the suffixes themselves

\item the ambiguity class assigned to each suffix

\end{itemize}

In the suffix calculation, the unique parameter that can be controlled
is their maximum lenght. It can be done by changing the value of the
variable {\tt class-guesser::*suffix-limit*}\footnote{This parameter has
been set to 5 by the authors.}.

The ambiguity class to be assigned to each suffix is selected from the
set of classes computed during normal training, which is written to a
classes file.  This file contains (i) every tag observed in the lexicon
(which is, obviously, unambiguous), (ii) every set of ambiguously
assigned tags for every form in the lexicon, and (iii) the ambiguity
class for the open class (a default class).

The above-mentioned function, after computing a suffix, observes words
in the lexicon ending in the proposed suffix and the set of tags
assigned to them. It then eliminates those tags not included in the
ambiguity class for the open class and, afterwards, tries to match the
remaining tags with one of the existing ambiguity classes. If it
succeeds, this ambiguity class is assigned to the suffix. Conversely,
if it fails, the suffix will receive the default ambiguity class.

While this behaviour may be correct for both non-inflecting languages
(as English) and relatively reduced tagsets, it is considered highly
inefficient for inflectional languages and more extensive tagsets. We
will try to exemplify this point in the following paragraph. \\

There are many ambiguous forms in the Spanish lexicon. Most cases range
over 2 to 4 tags for each form, but there are a few cases with even 5
or 6. If we establish an open class including all nominal,
adjectival, and verbal tags, the classes file will contain, along with
this open class, the list of individual tags of the tagset, the default
ambiguity class, several ambiguity classes formed by 2-tuples,
3-tuples, 4-tuples and a few 5-tuples and 6-tuples. This means that
computed suffixes must be accomodated into these latter ambiguity
classes in order to maximize accuracy in the assignment of tags (the
use of the default ambiguity class in these cases will produce
incorrect results in most cases).  Assuming that {\em a} is one of the
suffixes computed by the above-mentioned function, the problem then is
trying to match the set of tags observed in the lexicon for words
ending in {\em a} included in the intersection with the default
ambiguity class with one of the previously computed classes. Words
ending in {\em a} can, usually, be singular feminine adjectives or
nouns, subjunctive present first and third person singular verbs, and
indicative present third person singular verbs ({\tt \#(:ADJGFS :NCFS
:VLPI3S :VLPS1S :VLPS3S)})\footnote{Some masculine nouns and adjectives
can end in {\em a}, though only in a small number of cases.
Imperatives can also end in this suffix. However, these can be treated
as exceptions and included in the lexicon.}. Now, if we take wordforms
with 5 different tags, we learn that the number of classes generated by
these is limited to just four:

\begin{verbatim}
   #(:ADJGFS :ADJGMS :ADVGR :VLPPFS :VLPPMS)
   #(:ADJGFS :ADJGMS :NCMS :VLPPFS :VLPPMS)
   #(:ADJGFP :ADJGMP :NCMP :VLPPFP :VLPPMP)
   #(:ADJGFS :ADJGMS :PREP :VLPPFS :VLPPMS)
\end{verbatim}

Obviously, there is no possible matching between the former ambiguity
class and any of the latter. The former ambiguity class does simply not
exist ---there must exist at least one ambiguous form (ending in {\em
a} or in another suffix) validating an ambiguity class in order for it
to be selected when observed in words ending in {\em a}. The result
observed is that the function is forced to assign the open ambiguity
class to most of the suffixes computed.  \\

Moreover, in inflectional languages, the selection of the training
corpus is also crucial to the issue of suffix calculation. A sufficient
amount of text containing an as wide as possible range of words should
be gathered and used for training purposes. However, this prerequisite
alone does not guarantee a proper computation of suffixes, since the
function operates not only on word tokens from the training corpus but
also on the system's lexicon. The parameter to be considered in this
respect is not the actual size of this lexicon (which, nevertheless, is
important in order to accurately assign ambiguity classes to word
tokens from a corpus), but the set of ambiguity classes represented in
that lexicon ---and this set would not increase with the addition of
new words.

Likewise, inflectional languages, like Spanish, present the
characteristic of having a clear correspondance between (linguistically
motivated) suffixes and morphosyntactic properties of the word(s) they
are attached to. Consequently, this {\em a priori} knowledge could be
exploited in a tagging system like the one described here. Thus, if a
word ending in {\em a} can represent the following ambiguity class:

\begin{verbatim}
   "a" #(:ADJGFS :NCFS :VLPI3S :VLPS1S :VLPS3S)},
\end{verbatim}

\noindent
the system should be able to use this information without needing to
estimate it.

On the other hand, the practice of manual coding of information for
unknown words has been used only to a relative extent in probabilistic
models of language.  Some systems, like the Xerox Tagger, compute
probabilistically both the suffixes and the ambiguity classes
associated to them; but others, like the one described in
\cite{Weischedel93}, include a hybrid approach where suffixes are
manually added and ambiguity classes are approximated directly from
training data.

However, all probabilistic taggers work with manually coded information
as, for instance, a lexicon. Hence, a new approach could include both
manually-computed suffix tables and ambiguity classes, specially for
inflectional languages where this information can be straightforwardly
obtained, thus improving system accuracy. This approach, however, has
the drawback that migrating the system to a new tagset bears more
resource conversion work, since both the lexicon and the suffix table
will have to be mapped onto it. \\

In the light of this argumentation, a modification to the system has
been proposed and successfully implemented.  This consists in merging,
during normal training, the set of classes observed in the lexicon with
those stated by a linguist in the suffix file.  The training process
will benefit from the reduction in the number of elements of the
ambiguity classes to be computed when words not contained in the
lexicon are found, thus improving accuracy in the generation of paths.

The benifits of this methodology of work are shown in the following
sections.

\section{Model tuning}

Parameter estimation is a central issue in probabilistic models of
language.  A hidden Markov model of language can be tuned in a variety
of ways.  Thus, several decisions have been taken concerning the
tagset, the lexicon, and the {\em biases}. These choices are presented
and (hopefully) justified below. The selection of the training corpus
and the results obtained are also discussed.

\subsection{The tagset and the lexicon}

Tagsets used by taggers for English have been usually derived in some
way from that used in the Brown Corpus \cite{Francis82}, which
distinguishes 87 tags. The trend since the design of this tagset has
been to refine and elaborate it. Thus, the Lancaster-Oslo/Bergen (LOB)
Corpus distinguishes about 135 tags, and the Lancaster UCREL group uses
a set of 166 tags (for CLAWS2 \cite{Garside87}). Other tagsets are
even larger, as the one used in the London-Lund Corpus of Spoken English,
which contains 197 tags.

These further refinements of the original Brown tagset reflect the
necessity for a tagged corpus to show all the (morpho-)syntactic
idiosyncracies of a language. Thus, the rationale behind developing
large, richly articulated tagsets is to approach ``the ideal of
providing distinct codings for all classes of words having distinct
grammatical behaviour'' \cite[:167]{Sampson87}.

On the other hand, some projects based on a stochastic orientation have
modified the original Brown Corpus tagset by paring it down rather than
extending it. This is the case of the Penn Treebank Project, that uses
36 POS tags \cite{Marcus92}. The decision was founded not only on the
use of a probabilistic model but also on the fact that the goal was to
parse the corpus, thus some POS distinctions were recoverable with
reference to syntactic structure. \\

However, international initiatives on corpus annotation standards, as
those proposed by \mbox{EAGLES} \cite{MSAL21}, recommend the
distinction of major morphosyntactic categories within tagsets. In
fact, level 1 (L1), including {\em recommended attributes/values},
distinguishes, among others, {\em type}, {\em gender}, {\em number},
{\em case}, {\em person}, {\em tense}, {\em mood}, and {\em
finiteness}.  EAGLES recommendations explicitly state that ``[t]he
standard requirement for these {\em recommended} attributes/values is
that, if they occur in a particular language, then it is advisable that
the tagset of that language should encode them.'' \cite[:16]{MSAL21}

Consequently, in the construction of a tagset to be used by a
probabilistic tagger, a trade-off must be found between exhaustivity
and accuracy ---the more exhaustive the information encoded in the
tagset (the greater the tagset), the less accurate the tagging will be
(since the resulting model will be more complex and parameter
estimations less accurate).

This trade-off has been taken into account in the creation of the
tagset for Spanish to be used by the Xerox Tagger within this project.
In a first attempt, a quasi-{\em ideal} tagset was built, taking into
account not only EAGLES recommendations but also TEI guidelines on text
annotation \cite{AI1W3}, \cite{AI1W9}, \cite{AI1W2}. This {\bf full
tagset} is presented in \cite{TAGSTSP2}. It contains 479 POS tags
(there are also special tags for punctuation signs)\footnote{The final
version currently used is slightly different. It has 466 POS tags.}.
Thus, it is a very comprehensive tagset, distinguishing almost all
morphosyntactic features recommended by the abovementioned initiatives.
Some examples of information considered are presented below:

\begin{itemize}

\item Nouns:  {\em common/proper} distinction, with various subtypes for
propers; semantic information considered in the first tagset ({\em
temporal}, {\em locative}, {\em measurement}, {\em numeral}, and {\em
organization}) has been now restricted only to {\em measurement}, given
the large amount of postediting derived from the initial distinction;
other common morphosyntactic information ({\em gender}, {\em number}).

\item Adverbs: {\em degree}, {\em wh information}, {\em locative} (with
subtypes), {\em deixis}, and {\em polarity}.

\item Verbs:  {\em status} (main/auxiliaries), {\em person}, {\em
number}, {\em tense}, {\em mood}, {\em gender}, and {\em finiteness}
(implicit). Given the rich verbal morphology of Spanish, verbal tags
account for 59\% of the total number of tags.

\end{itemize}

This tagset has been considered ``too finegrained to be suitable for a
probabilistic tagger'' [Lauri Karttunen, personal communication]. \\

Then, a second, {\bf reduced tagset}, based on the first one, was
built.  The number of tags has been dramatically cut down in this
tagset to 174.  Features previously considered for major categories
have been restricted to {\em gender}, {\em number} and {\em
person}\footnote{Besides, status of verbs has also been taken into
account. Semantic information on nouns has been eliminated, though
proper names and names of the days of the week and of the months have
specific tags.}. Minor categories have also seen reduced the
morphosyntactic (and sometimes semantic) information considered at
first.

The reduced tagset has been built precisely with the idea of testing the
improvement of the tagging accuracy when the number of parameters
is simpler.  \\

All probabilistic taggers make use of a lexicon of varied coverage.
\cite{Cutting92}, for instance, report on tagging results on even
numbered sentences of the Brown corpus using a 50,000 forms lexicon.
With this lexicon and the suffix file, no unknown forms were
encountered in the training process, thus providing no training data for
forms assigned the open class.

However, a greater lexicon does not necessarily guarantee a better
tagging accuracy. Words are usually ambiguous and may take, depending
on the context, a different POS tag. The probability of a given word
taking one or the other tag may not be the same, though, and some
systems have the possible tags for a word arranged in a decreasing
likelihood, and also include special mechanisms to express the fact
that certain tags are ``rare'' or ``very rare'' \cite{Garside87}.  When
this selection is impossible in the system, other resorts may be
employed to reduce ambiguity. Some authors use an optimal dictionary
that indicates, for each word, all the tags assigned to it somewhere in
the corpus being used, but not other, possible tags \cite{Merialdo94}.
Others propose the exclusion of rare readings from the lexicon to
prevent the tagger from selecting them \cite{Tapanainen94}.

Since our starting point is not a tagged corpus in which to perform the
testing of a given stochastic model, our lexicon is not specially biased
towards the corpus we aim at tagging. On the contrary, we would like to
build the tagger on as a uniform lexical material as possible. Hence,
the whole set of tags for each word has been taken into account during
lexicon building.

The lexicon used by the system has been produced by compiling different
sources of information, although some coding work has also been
performed. This lexicon is being used in the actual tagging of the ITU
corpus, since it provides a more accurate model to lexical ambiguity
than that provided by suffix information alone\footnote{Nevertheless,
this lexicon has to be used carefully.  The sources for lexical
information are free from error. In fact, morphosyntactic information
has been observed to be wrong in some cases. An overall correction of
the lexicon is being carried out.}.

\subsection{Training a hidden Markov model}

Training on hidden Markov models of language is performed without a
tagged corpus. In a tagger under this regime, state transitions (i.e.,
transitions between categories) are unobservable. Under these
circumstances, the training is performed according to a Maximum
Likelihood principle, using the Forward-Backward (FB) or Baum-Welch
algorithm. This training process can be biased in a number of ways in
order to `force' somehow the learning process. Two such ways
implemented in the Xerox Tagger, concerning ambiguity classes and state
transitions, are described below:

\begin{itemize}

\item The biasing facts on ambiguity classes are called {\em symbol
biases}. These represent a kind of lexical probabilities for given
equivalence classes. This way, ambiguity classes are annotated with
favoured tags. Note, however, that this is stated for a given class and
not for individual forms in the lexicon (as it is, for instance, in
CLAWS \cite{Garside87}), resulting in a less efficient mechanism.

\item The biasing facts on state transitions are called {\em transition
biases}. These specify that it is likely or unlikely that a tag is
followed by some specific tag(s). The biasing can be formulated either
as favoured or as disfavoured probabilities. Disfavoured probabilities
receive a small constant but are not disallowed; on the contrary, data
in the training corpus may modify probabilities.

\end{itemize}

\cite{Tapanainen94}, who use the Xerox Tagger in combination with ENGCG
for tagging English texts, reporting an accuracy of 98.5\%, propose
other ways of tuning the system. These are the following:

\begin{itemize}

\item Not including rare readings in the lexicon in order for the tagger
not to select them.

\item Using different values for the number of iterations (the number of
times the same block is used in training) and the size of the block of
text used for training.

\item The choice of the training corpus affects the result.

\end{itemize}

In our case, it has already been commented that an {\em a priori}
decision was testing the system with no special lexical limitations,
that is, with the whole set of possible tags for each word assigned to
it when included in the lexicon. With regard to the second suggestion,
initial parameters proposed by Xerox Tagger developers have been
preserved in order not to introduce more complexity to the initial
parameter estimation. Finally, the choice of the training corpus has
consequences on the accuracy of the system. As demonstrated by
\cite{Merialdo94}, when using an HMM, a greater training corpus does not
necessarily guarantee a better accuracy. On the contrary, an initial
model estimated by performing Relative Frequency (RF) training on a
tagged text may degrade if a relatively great untagged corpus is used
next. We don't have the possibility to perform a combined (RF and ML)
training but, in any case, the potential degradation of the model has
been taken into account when producing the final model. \\

The model has been initially tuned by means of the addition of both
transition and symbol biases. These have not been documented yet, but
they include favouring clitic-verb, determiner-noun and noun-adjective
transitions, and disfavouring adjective-adjective and
preposition-finite verb transitions. Nouns are favoured when they can
also be adjectives. \\

\subsection{Training corpus and results}

The system has been trained using both versions of the tagset.
Although the decision of tagging the ITU corpus using the {\em full}
version of the tagset was already adopted and postediting on the corpus
so tagged has begun, a parallel development of the tagger with the {\em
reduced} tagset has been performed.  Results obtained with both tagset
are presented in this section.  \\

The full 1M word subset of the corpus being postedited has been used as
the training corpus, leaving file {\tt SP\_itu\_corpus\_000} as the test
corpus.  This corpus contains 9,366 tagged tokens. The corpus has been
used in an incremental way, testing results with each partial model
obtained.  \\

In both cases, the system used includes an initial set of transition and
symbol biases which is responsible for the good results obtained with
the uniform (untrained) model.  The biasing facts are the same for each
model, as well as the lexicon (in terms of coverage) and the suffix
information file. \\

As it will become clear by looking at the results, there is not a clear
learning curve.  The system performs relatively well with the set of
initial biases, and even its accuracy improves 2.5\% with a small amount
of text.  However, best results are obtained with as less as 50,000
words, being the accuracy from this corpus size on almost the same.
Anyway, since results with other models are so close, it could be
difficult to prove Merialdo's claim.  \\

With respect to the comparison between both tagsets, the curve is the
same in either case, having also obtained the best results with the same
amount of training text. Surprisingly, the accuracy is also the same for
both tagsets with the best model. However, in general, the {\em reduced}
tagset shows an insignificant .1\% better accuracy than the {\em full}
tagset\footnote{All results reported refer to version 1.2 of the Xerox
Tagger. With version 1.1, the tagging produced is always the same
irrespective of the training corpus used. Not surprisingly, the HMM
file is also the same and the training times are suspiciously short.
Thus, version 1.1 seems to learn nothing from the training corpus.}. \\

Table~1 shows the behaviour of the system when tagging the test corpus
with the {\em full} tagset.

\vspace{.4cm}

\tiny
\begin{table}[htbp]
\caption{Statistics for the training using the {\em full} tagset.}

\begin{minipage}{16cm}
\begin{tabular}{||p{3cm}|p{2cm}|p{2cm}|p{2.5cm}|p{2cm}||} \hline
Training files & Word count\footnote{As counted by {\sc unix}
command {\tt wc}} & Training time\footnote{Real time} & Errors tagging
test corpus\footnote{First figures represent absolute number of errors;
second figures do not include foreign words} & Accuracy \\ \hline
No training & 0 & --            & 1059 - 645    & 88.69 - 93.11 \\
001-003 & 29931 & 30'24''       & 819 - 405     & 91.26 - 95.68 \\
001-006 & 53300 & 53'55''       & 790 - 376     & 91.51 - 95.93 \\
001-009 & 66922 & 1h 06'48''    & 855 - 441     & 90.87 - 95.29 \\
001-014 & 96603 & 1h 37'01''    & 840 - 426     & 91.03 - 95.45 \\
001-019 & 143129 & 2h 23'14''   & 853 - 439     & 90.89 - 95.31 \\
001-024 & 180302 & 2h 59'47''   & 830 - 416     & 91.14 - 95.56 \\
001-029 & 213518 & 3h 27'55''   & 827 - 413     & 91.17 - 95.59 \\
001-034 & 255960 & 4h 21'03''   & 835 - 421     & 91.08 - 95.50 \\
001-039 & 293203 & 4h 52'52''   & 833 - 419     & 91.11 - 95.53 \\
001-044 & 333570 & 5h 26'52''   & 832 - 418     & 91.12 - 95.54 \\
001-049 & 371338 & 6h 05'20''   & 835 - 421     & 91.08 - 95.50 \\
001-054 & 401433 & 6h 38'39''   & 833 - 419     & 91.11 - 95.53 \\
001-059 & 424189 & 6h 58'21''   & 832 - 418     & 91.12 - 95.54 \\
001-064 & 427487 & Out of heap  & --            & -- \\
001-069 & 507608 & 8h 16'51''   & 829 - 415     & 91.14 - 95.56 \\
001-074 & 586608 & 9h 36'16''   & 828 - 414     & 91.16 - 95.58 \\
001-079 & 637565 & 10h 15'57''  & 835 - 421     & 91.08 - 95.50 \\
001-084 & 698788 & 11h 13'50''  & 834 - 420     & 91.10 - 95.52 \\
001-089 & 776407 & 12h 25'55''  & 829 - 415     & 91.14 - 95.56 \\
001-094 & 823498 & 13h 20'34''  & 827 - 413     & 91.17 - 95.59 \\
001-099 & 890247 & 14h 16'14''  & 825 - 411     & 91.19 - 95.61 \\
001-106 & 971163 & 15h 47'20''  & 832 - 418     & 91.12 - 95.54 \\ \hline
\end{tabular}
\end{minipage}
\end{table}

\normalsize
\vspace{.5cm}

\newpage

Table~2 shows the behaviour of the system when tagging the test corpus
with the {\em reduced} tagset.

\vspace{.3cm}

\tiny
\begin{table}[htbp]
\caption{Statistics for the training using the {\em reduced} tagset.}

\begin{minipage}{16.4cm}
\begin{tabular}{||p{3cm}|p{2cm}|p{2cm}|p{2.5cm}|p{2cm}||} \hline
Training files & Word count\footnote{As counted by {\sc unix}
command {\tt wc}} & Training time\footnote{Real time} & Errors tagging
test corpus\footnote{First figures represent absolute number of errors;
second figures do not include foreign words} & Accuracy \\ \hline
No training & 0 & --            & 1032 - 618    & 88.98 - 93.40 \\
001-003 & 29931 & 20'12''       & 804 - 390     & 91.42 - 95.84 \\
001-006 & 53300 & 37'02''       & 790 - 376     & 91.51 - 95.93 \\
001-009 & 66922 & 46'00''       & 848 - 434     & 90.95 - 95.37 \\
001-014 & 96603 & 1h 09'43''    & 836 - 422     & 91.07 - 95.49 \\
001-019 & 143129 & 1h 40'30''   & 827 - 413     & 91.17 - 95.59 \\
001-024 & 180302 & 2h 05'14''   & 825 - 411     & 91.19 - 95.61 \\
001-029 & 213518 & 2h 25'07''   & 819 - 405     & 91.26 - 95.68 \\
001-034 & 255960 & 2h 56'57''   & 822 - 408     & 91.22 - 95.64 \\
001-039 & 293203 & 3h 19'50''   & 837 - 423     & 91.06 - 95.48 \\
001-044 & 333570 & 3h 47'51''   & 832 - 418     & 91.12 - 95.54 \\
001-049 & 371338 & 4h 20'24''   & 831 - 417     & 91.13 - 95.55 \\
001-054 & 401433 & 4h 35'15''   & 823 - 409     & 91.21 - 95.63 \\
001-059 & 424189 & 4h 48'16''   & 820 - 406     & 91.24 - 95.66 \\
001-064 & 427487 & 4h 51'21''   & 820 - 406     & 91.24 - 95.66 \\
001-069 & 507608 & 5h 45'26''   & 818 - 404     & 91.27 - 95.69 \\
001-074 & 586608 & 6h 41'45''   & 818 - 404     & 91.27 - 95.69 \\
001-079 & 637565 & 7h 12'32''   & 818 - 404     & 91.27 - 95.69 \\
001-084 & 698788 & 7h 50'53''   & 813 - 399     & 91.32 - 95.74 \\
001-089 & 776407 & 8h 40'21''   & 816 - 402     & 91.29 - 95.71 \\
001-094 & 823498 & 9h 13'37''   & 812 - 398     & 91.33 - 95.75 \\
001-099 & 890247 & 9h 54'11''   & 817 - 403     & 91.28 - 95.70 \\
001-106 & 971163 & 10h 54'24''  & 821 - 407     & 91.23 - 95.65 \\ \hline
\end{tabular}
\end{minipage}
\end{table}

\normalsize
\vspace{.4cm}

\section{Benefits of a linguistically enriched model}

Apart from the reasons mentioned in previous sections relative to
the soundness of a model based on linguistic knowledge in order to
treat suffix information, at least for inflectional languages, there is
also a kind of pragmatic reason:  tagging should be more accurate using
a linguistically enriched model than with the original, only
statistical one. In order to prove this statement, a comparison of the
performance of both models will be carried out. For the moment, suffix
information files can be compared in order to guess which the best model
will be.

The whole subset of the corpus to be postedited for alignment within
the CRATER project has been considered as the training corpus for the
function that computes suffixes. Results obtained both by hand and
automatically are presented in table~3:

\vspace{.4cm}
\tiny

\begin{table}[htbp]
\caption{Suffix file information.}

\begin{minipage}{16.4cm}
\begin{tabular}{||p{4cm}||p{.7cm}|p{.7cm}|p{.7cm}|p{.7cm}|p{.7cm}|p{.7cm}|p{.7cm}|p{.7cm}|p{.7cm}|p{.7cm}||} \hline
Model & \multicolumn{2}{c}{Manually added} &
        \multicolumn{8}{|c||}{Automatically computed} \\ \hline
Tagset & \multicolumn{1}{|c|}{full} & \multicolumn{1}{c}{reduced}  &
\multicolumn{4}{|c}{full} & \multicolumn{4}{|c||}{reduced} \\ \hline
Previously trained model & -- & -- & \multicolumn{2}{|c|}{no} &
        \multicolumn{2}{|c|}{yes} & \multicolumn{2}{|c|}{no} &
        \multicolumn{2}{|c||}{yes} \\ \hline
Maximum suffix lenght \mbox{parameter} & -- & -- & 15 & 5 & 15 &
5 & 15 & 5 & 15 & 5 \\ \hline
Number of suffixes & 208\footnote{Besides, this file includes 306
suffixes for the recognition of verbs with enclitics and 22 suffixes
for foreign words.} & 208\footnote{Besides, this file includes 306
suffixes for the recognition of verbs with enclitics and 22 suffixes
for foreign words.} & 16 & 94 & 16 & 100 & 16 & 78 & 16 & 77 \\
Maximum suffix lenght & -- & -- & 1 & 4 & 1 & 4 & 1 & 4 & 2 & 4 \\
Total number of tags & 376 & 362 & 97 & 445 & 97 & 340 & 87 & 418 & 51 & 311 \\
Tags per suffix & 1.8 & 1.7 & 6 & 4.7 & 6 & 3.4 & 5.4 & 5.4 & 3.2 & 4.1 \\
\hline
\end{tabular}
\end{minipage}
\end{table}

\vspace{.5cm}

\normalsize

Note that the function automatically calculating suffix information can
be executed both with a trained and with an untrained model. Results,
however, are better with a previously trained model. Nontheless, these
results are far from those obtained with the manually included
information. Besides, the number of suffixes is smaller and the maximum
lenght does not guarantee the recognition of typical unambiguous
suffixes:  {\em -mente}, which is always an adverb, or {\em -ci\'on},
always a feminine singular noun.\footnote{The corpus being tagged has
been converted to a shallow SGML representation, specially concerning
8-bit characters. Note that the SGML representation of ISO LATIN
characters converts the latter suffix into {\em -ci\&oacute;n},
resulting, then, impossible its identification with a suffix limit of 5
characters long.} Other major drawback of the function is that does not
take account of case of words, hence producing suffixes in uppercase
and/or lowercase with different information in either case. \\

Consequently, the performance of a model using the approach proposed
will be better for Spanish than the original strategy of the tagger.

\section{Other issues}

The Xerox Tagger lacks the adequate mechanisms for the treatment of
complex lexical elements. Segmentation of the text into tokens is
performed by means of graphic information like space characters and
other delimiters. This poses a problem for the identification of both
complex orthographic words comprising more than one textword (i.e.
clitic forms, since portmanteaux are to be assigned a specific tag) and
textwords that span over more than one orthographic word (i.e.
continuous invariant multi-word units).

The first issue, still in the implementation phase, includes the
segmentation of higher-order complex words for the recognition and
further tagging of verbal forms with enclitics. So far, the system
assigns a special tag, {\tt VCLI}, to these forms. Nevertheless, the
code is being changed so as to split these tokens and appropriately tag
these elements. The complexity of this task stresses somehow one of the
limitations of the Xerox Tagger ---the system lacks a morphological
analyzer. This limitation questions one of the claims of the
developers, namely, its language-indepency. The lexical repertoire of
fully inflected forms in languages a high inflectional productivity may
collapse the system. This is also true for highly agglutinative
languages, where productive word-formation rules make impossible the
creation of a wide-coverage lexicon.

The second issue is has been solved by means of a pre-processing
phase.  Space characters separating components of a complex textword
are replaced by an underscore character (\_), thus normal tokenization
may operate on these items. To this end, a program developed by
Theo~W.~Tams, from {\sc Eurotra-DK} \cite{Jensen90}, during the third
phase of \mbox{\sc Eurotra-I} for a tender on Front End Integration has
been used. The source code has been adapted to our requirements. \\

Along with these, other modifications have been performed to the
original Xerox Tagger. Thus, the ouput format has been modified, so that
instead of being presented in the following line, tags are placed to the
right of every word separated from it by means of an underscore
character (\_), as it is usual in the taggers from England, specially
in the works by the University of Lancaster.

Sentence boundaries are correctly identified by the tokenizer, with
the usual limitations inherent to the tasks ---as, for instance, the
proper distinction between these and dots in abbreviations. This issue
is esential to the system behaviour given that the training process is
performed on text chunks that are segmented into sentences.

However, since the full stop is used for the purpose of sentence
identification it cannot be properly tagged by the tokenizer itself. An
automatic postediting phase, that currently performs also the correction
of certain transitions between categories that the system tends to tag
incorrectly, carries out the correct tagging of full stops.

Besides, special tokenization rules have been implemented in order to
recognize two date formats, as observed in the ITU corpus, namely, {\tt
dd.mm.yy} and {\tt yyyy-yyyy}. \\

\section{Conclusions}

This paper presents results obtained with the port to Spanish of a
public domain tagger ---the Xerox Tagger. With some modifications,
necessary in our view, to tag inflectional languages (treatment of the
suffix information for the guesser) and for the proper segmentation of
complex words (verbal forms with ewclitics), the system behaves with the
usual error rates accepted for other morphosyntactic taggers. So far,
the model has been tested with free text, but only with the ITU corpus.
Results may be poor in this case. \cite{Chanod95} present, using a
corpus different to that used for training, results sensibly better for
French (96.8\% of accuracy) than those presented here for Spanish, while
the accuracy for German is 96.66\% \cite{Feldweg95}. Nevertheless, the
great difference these two tagsets with respect to the one used in
CRATER must be taken into account. While \cite{Chanod95} use a tagset
consisting of 88 tags and \cite{Feldweg95} a tagset with 42 tags, our
tagset has 466 different tags. These tagset size may be excesive,
specially  for a probabilistic tagger, but results obtained with our
corpus show similar accuracy, with the value-added benefit for the
tagged corpus of having the whole variety of morphosyntactic categories
and subcategories reflected in it\footnote{The Spanish version of this
tagger will be in the public domain in october, 1995.}.

\vspace{5mm}

\small
\noindent
{\bf Acknowlegments}

\noindent
We are grateful to Flora Ram\'{\i}rez Bustamante for her cooments and
help in the building of the linguistic resources on which this tagger
has been developed. Ruthanna Barnett has also provided her grain of sand
with her comments.


\begin{thebibliography}{99}

\bibitem[Nieto, 1994]{Anieto94}A. F. Nieto. {\em CRATER: UPM Progress
for the Period April-September 1994}. CRATER Internal Document.
September 1994.

\bibitem[Cutting {\it et al.}, 1992]{Cutting92}D. Cutting, J. Kupiec,
J. Pedersen, and P. Sibun. A Practical Part-of-Speech Tagger.  In {\em
Proceedings of the Third Conference on Applied Natural Language
Processing}, Trento.

\bibitem[Chanod and Tapanainen, 1995]{Chanod95}J.-P. Chanod and P.
Tapanainen.Tagging French -- comparing a statistical and a
constraint-based method. In {\em Proceedings of the EACL-95}, Dublin.

\bibitem[Derouault and Merialdo, 1986]{Derouault86}A. M. Derouault and
B. Merialdo. Natural Language Modelling for Phoneme-to-Text
Transcription. {\em IEEE Transactions on Pattern Analysis and Machine
Intelligence}, PAMI-8:742--749.

\bibitem[Feldweg, 1995]{Feldweg95}H. Feldweg. Implementation and
evaluation of a German HMM for POS disambiguation. In {\em Proceedings
of the EACL SIGDAT workshop 1995}, Dublin.

\bibitem[Francis and Ku\v{c}era, 1982]{Francis82}W. N. Francis and H.
Ku\v{c}era. {\em Frequency analysis of English usage. Lexicon and
grammar}, Houghton Mifflin, Boston.

\bibitem[Garside {\it et al.}, 1987]{Garside87}R. Garside, G. Leech and
G.  Sampson.  {\em The Computational Analysis of English.  A
Corpus-Based Approach}, Longman, London.

\bibitem[Jelinek, 1985]{Jelinek85}F. Jelinek. Markov Source Modeling of
Text Generation. In J. K. Skwirzinski, editor, {\em Impact of Processing
Techniques on Communication}, Nijhoff, Dordrecht.

\bibitem[Jensen {\it et al.}, 1990]{Jensen90}N. Jensen, T. Tams, N. Jaeger
and V. Pirrelli. {\em Final Report on Front End Integration}. {\sc
Eurotra} Internal Document.

\bibitem[Karlsson {\it et al.}, 1994]{Karlsson94}F. Karlsson, A.
Voutilainen, J. Heikkil\"{a} and A. Anttila (eds.) {\em Constraint
Grammar: a Language-Independent System for Parsing Unrestricted Text}.
Mouton de Gruyter, Berlin.

\bibitem[Kupiec, 1989]{Kupiec89b}J. M. Kupiec. Probabilistic Models of
Short and Long Distance Word Dependencies in Running Texts. In {\em
Proceedings of the 1989 DARPA Speech and Natural Language Workshop},
pages 290--295, Morgan Kaufman, Philadelphia.

\bibitem[Langendoen and Fahmy, 1991]{AI1W9}D. T. Langendoen and E.
Fahmy.  {\em Feature-structure markup for presentation at Oxford and
Brown workshops}, Department of Linguistics, University of Arizona,
Tucson, AZ 85721 USA, September.

\bibitem[Leech and Wilson, 1994]{MSAL21} G. Leech and A.  Wilson.  {\em
Draft Sections 4.6 and 4.7 ofthe EAGLES Interim Report:  Annotation
Sub-Group}, EAGLES, February.

\bibitem[Marcus and Santorini, 1992]{Marcus92}M. P. Marcus and B.
Santorini. {\em Building very large natural language corpora: the
Penn Treebank}, Department of Computer and Information Science,
University of Pennsylvania, Philadelphia, PA 19104, January.

\bibitem[Merialdo, 1994]{Merialdo94}B. Merialdo. Tagging English Text
with a Probabilistic Model. {\em Computational Linguistics}, 20(2),
155--171.

\bibitem[Sampson, 1987]{Sampson87}G. Sampson. Alternative grammatical
coding systems. In Garside {\it et al.}. {\em The Computational Analysis
of English. A Corpus-Based Approach}, Longman, London, 165--183.

\bibitem[S\'anchez-Le\'on, 1994]{TAGSTSP2}F. S\'anchez Le\'on.  {\em
Spanish tagset for the CRATER project}, CRATER Internal Document, March.
Also available through WWW as http://xxx.lanl.gov/cmp-lg/9406023.

\bibitem[Simons, 1991]{AI1W3}G. F. Simons. {\em Feature System
Declarations and the Interpretation of Feature Structures}, January
1991.

\bibitem[Tapanainen and Voutilainen, 1994]{Tapanainen94}P. Tapanainen
and A. Voutilainen. Tagging accurately -- Don't guess if you know. To
appear in {\em Proceedings of the Fourth Conference on Applied Natural
Language Processing}, Stuttgart.

\bibitem[TEI AI1W2, 1991]{AI1W2}Text Encoding Initiative.  {\em TEI AI
1W2.  List of Common Morphological Features For Inclusion in TEI
Starter Set Of Grammatical-Annotation Tags}, June.

\bibitem[Weischedel {\it et al.}, 1993]{Weischedel93}R. Weischedel, M.
Meteer, R. Schwartz, L. Ramshaw, and J. Palmucci. Coping with Ambiguity
and Unknown Words through Probabilistic Models. {\em Computational
Linguistics}, 19(2), 359--382.

\end{thebibliography}
\end{document}